\documentclass[letter, 10pt, journal,twoside]{ieeetran}      % Use this line for a4 paper

\IEEEoverridecommandlockouts                              % This command is only needed if 
\usepackage{graphicx,epstopdf}
\usepackage{amsmath}
\usepackage{amssymb}
\usepackage{url}

\usepackage{color, colortbl}

\usepackage{array,multirow,graphicx}

\newcommand{\argmin}{\operatornamewithlimits{argmin}}
\graphicspath{ {imgs/} }
\usepackage{tikz}

\usepackage[markup=bfit, deletedmarkup=sout, authormarkup=brackets]{changes}
\definecolor{Gray}{gray}{0.9}
\definecolor{Gray2}{rgb}{0.1,0.1,0.1}
\usepackage{draftwatermark}

\title{Robot self-calibration using multiple kinematic chains -- a simulation study on the iCub humanoid robot}
\author{Karla Stepanova$^{1,2}$, Tomas Pajdla$^{2}$, and Matej Hoffmann$^{1}$% <-this % stops a space
\thanks{Manuscript received: September 10, 2018; Revised: December 11, 2018; Accepted: January 16, 2019.}
\thanks{This paper was recommended for publication by Editor Nikos Tsagarakis upon evaluation of the Associate Editor and Reviewers'
comments. K.S. and M.H. were supported by the Czech Science Foundation under Project GA17-15697Y; T.P. was supported by the European Regional Development Fund under the project Robotics for Industry 4.0 (reg. no. CZ.02.1.01/0.0/0.0/15 003/0000470).}
\thanks{$^{1}$Karla Stepanova and Matej Hoffmann are with Department of Cybernetics, Faculty of Electrical Engineering, Czech Technical University in Prague, Czech Republic. {\tt\footnotesize  karla.stepanova@cvut.cz} and {\tt\footnotesize matej.hoffmann@fel.cvut.cz}.}%
\thanks{$^{2}$Karla Stepanova and Tomas Pajdla are with Czech Institute of Informatics, Robotics, and Cybernetics, Czech Technical University in Prague, Czech Republic	{\tt\footnotesize  { pajdla@cvut.cz}}.}%
\thanks{Digital Object Identifier (DOI): see top of this page.}

}

\begin{document}
\SetWatermarkAngle{0}
\SetWatermarkColor{black}
\SetWatermarkLightness{0.5}
\SetWatermarkFontSize{10pt}
% \SetWatermarkScale
% \SetWatermarkHorCenter
\SetWatermarkVerCenter{20pt}
\SetWatermarkText{\parbox{30cm}{%
\centering This is the author final version of the manuscript published in IEEE Robotics and Automation Letters 4(2), pp. 1900-1907\\
\centering https://doi.org/10.1109/LRA.2019.2898320, (C) IEEE.}}

\maketitle

%%%%%%%%%%%%%%%%%%%%%%%%%%%%%%%%%%%%%%%%%%%%%%%%%%%%%%%%%%%%%%%%%%%%%%%%%%%%%%%%
\begin{abstract}
Mechanism calibration is an important and non-trivial task in robotics. Advances in sensor technology make affordable but increasingly accurate devices such as cameras and tactile sensors available, making it possible to perform automated self-contained calibration relying on redundant information in these sensory streams. In this work, we use a simulated iCub humanoid robot with a stereo camera system and end-effector contact emulation to quantitatively compare the performance of kinematic calibration by employing different combinations of intersecting kinematic chains---either through self-observation or self-touch. The parameters varied were: (i) type and number of intersecting kinematic chains used for calibration, (ii) parameters and chains subject to optimization, (iii) amount of initial perturbation of kinematic parameters, (iv) number of poses/configurations used for optimization, (v) amount of measurement noise in end-effector positions / cameras. The main findings are: (1) calibrating parameters of a single chain (e.g. one arm) by employing multiple kinematic chains (``self-observation'' and ``self-touch'') is superior in terms of optimization results as well as observability; (2) when using multi-chain calibration, fewer poses suffice to get similar performance compared to when for example only observation from a single camera is used; (3) parameters of all chains (here 86 DH parameters) can be subject to calibration simultaneously and with 50 (100) poses, end-effector error of around 2 (1) mm can be achieved; (4) adding noise to a sensory modality degrades performance of all calibrations employing the chains relying on this information.
\end{abstract}


\section{INTRODUCTION}

\IEEEPARstart{R}{obots} performing manipulation tasks rely on models of their bodies and their success is largely determined by their accuracy. However, inaccuracies creep in many ways as for example in the assembly process, in mechanical elasticity, or simply because of cheap design of components. Therefore, the actual model parameters of every robot exemplar have to be found by means of a calibration procedure, usually relying on external metrology systems. For kinematic calibration, such apparatuses can measure one or more of the components of the end-effector pose employing mechanical, visual, or laser systems (see~\cite{Hollerbach2016} for a survey). Different arrangements have different accuracy, requirements on the environment, and cost. These conditions have to be present for recalibration to be performed. 

Current trends in the robotics industry make classical calibration procedures less practical: with the advent of the so-called ``collaborative robots'', for example, the machines are becoming cheaper, lightweight, compliant, and they are being deployed in more versatile ways according to the needs of customized production of smaller batches rather than being fixed in a single production line for their entire lifetime. All these factors increase the need for calibration to be performed more frequently. At the same time, the machines, including home and service robots, often come with richer sets of powerful sensory devices that are affordable and not difficult to operate. Both these trends speak for alternative solutions to the self-calibration problem that are more ``self-contained'' and can be performed autonomously by the robot.

Hollerbach et al.~\cite{Hollerbach2016} classify different calibration methods into \textit{open-loop}---where one or more of the components of the end-effector pose is measured employing mechanical, visual, or laser systems---and \textit{closed-loop} where physical constraints on the end-effector position or orientation can substitute for measurements. Observing the end-effector---or in general any other points on the kinematic chain---using a camera falls into the open-loop calibration family, although components of the end-effector pose can be observed only indirectly through projection into the camera frame. Self-touch configurations employing two arms of the humanoid robot could be framed as a constraint if contact measurement only (e.g. from force/torque sensors) was available and hence treated as closed-loop. In this work, we follow up on \cite{Roncone_ICRA_2014} and emulate sensitive skin measurements, which provide the position of contact (and hence fit more naturally with open-loop calibration).

Our work is a simulation study that draws on calibration in the real world---like different approaches to kinematic calibration of the iCub humanoid robot relying on self-observation \cite{Fanello2014,Vicente2016} and self-touch \cite{Roncone_ICRA_2014}.
Using the model of the robot with identical parameters, but exploiting the fact that we have complete knowledge of the system and capacity to emulate different levels of model perturbation and measurement noise, our goal is to get insights into the pros and cons of different optimization problem formulations. In particular, we study how the calibration performance is dependent on the type and number of intersecting kinematic chains, the number of parameters calibrated, number of robot configurations, and the measurement noise. Accompanying video is available here \url{https://youtu.be/zP3c7Eq8yVk} and dataset at~\cite{ProjectWeb}.

This article is structured as follows. Related work is reviewed in the next section, followed by Materials and Methods, Data Acquisition and Description, and Simulation Results. We close with a Discussion and Conclusion. 

\section{RELATED WORK}
We focus on humanoid robots or humanoid-like setups with many Degrees of Freedom (DoF) of two arms that can possibly self-touch, equipped with cameras and  tactile or inertial sensors. These are challenging setups for calibration but they create new opportunities for automated self-contained calibration based on closing kinematic loops by touch (self-contact) and vision.

Most often, the loops are closed through self-observation of the end-effector using cameras located in the robot head (\textit{open-loop calibration} method per \cite{Hollerbach2016}). Hersch et al. \cite{Hersch2008} and Martinez-Cantin et al.~\cite{Martinez-Cantin2010} present online methods to calibrate humanoid torso kinematics relying on gradient descent and recursive least squares estimation, respectively. The iCub humanoid was employed in \cite{Fanello2014,Vicente2016}. Vicente et al.~\cite{Vicente2016} used a model of the hand's appearance to estimate its 6D pose and used that information to calibrate the joint offsets. Fanello et al.~\cite{Fanello2014} had the robot observe its fingertip and learned essentially a single transformation only to account for the discrepancy between forward kinematics of the arm and the projection of the finger into the cameras. 

Next to cameras, inertial sensors also contain information that can be exploited for calibration. Kinematic calibration was shown exploiting 3-axis accelerometers embedded in the artificial skin modules distributed on robot body \cite{Mittendorfer2012,Dean2018} or in the control boards on the iCub \cite{Guedelha2016} or CMU/Sarcos \cite{Yamane2011}.

The advent of robotic skin technologies \cite{Bartolozzi2016,Dahiya2013} opens up the possibility of a new family of approaches, whereby the chain is closed through contact like in closed-loop calibration, but the contact position can be extracted from the tactile array. Roncone et al.~\cite{Roncone_ICRA_2014} showed this on the iCub robot that performs autonomous self-touch using a finger with sensitive fingertip to touch the skin-equipped forearm of the contralateral arm; Li et al.~\cite{QiangLi2015} employed a dual KUKA arm setup with a sensorized ``finger'' and a tactile array on the other manipulator. Forward kinematics together with skin calibration provide contact position that can then be used for robot kinematic calibration. In this sense, the skin provides a pose measurement rather than constraint and as such, this may fall under \textit{open-loop calibration}. In this way, one arm of a humanoid can be used to calibrate the other. Khusainov et al.~\cite{Khusainov2017} exploit this principle using an industrial manipulator to calibrate the legs of a humanoid robot. Another variant is exploiting the sensitive fingertips to touch a known external surface \cite{Zenha2018}.

Birbach et al. ~\cite{Birbach2015} were to our knowledge the only ones to employ truly ``multisensorial'' or ``multimodal'' calibration. Using the humanoid robot Justin observing its wrist, the error functions comparing the wrist's position from forward kinematics with its projection into the left and right camera images, Kinect image, and Kinect disparity, together with an inertial term, were aggregated into a single cost function to be minimized. It is claimed that while pair-wise calibration can lead to inconsistencies, calibrating everything together  in a ``mutually supportive way'' is most efficient. 

In this work, we compare calibration through self-observation (with projection into cameras) and calibration through self-touch and the effect of their synergy. Our work makes a unique contribution, also compared to \cite{Birbach2015} who, first, employ essentially only ``hand-eye'' kinematic chains terminating in different vision-like sensors in the robot head, and, second, consider only the case where all chains are combined together using a single cost function.

\section{MATERIALS AND METHODS}
\subsection{iCub robot kinematic model and camera parameters}
\begin{figure}[thpb]
\centering
      \framebox{\parbox{3.2in}{ \includegraphics[width = 95 pt]{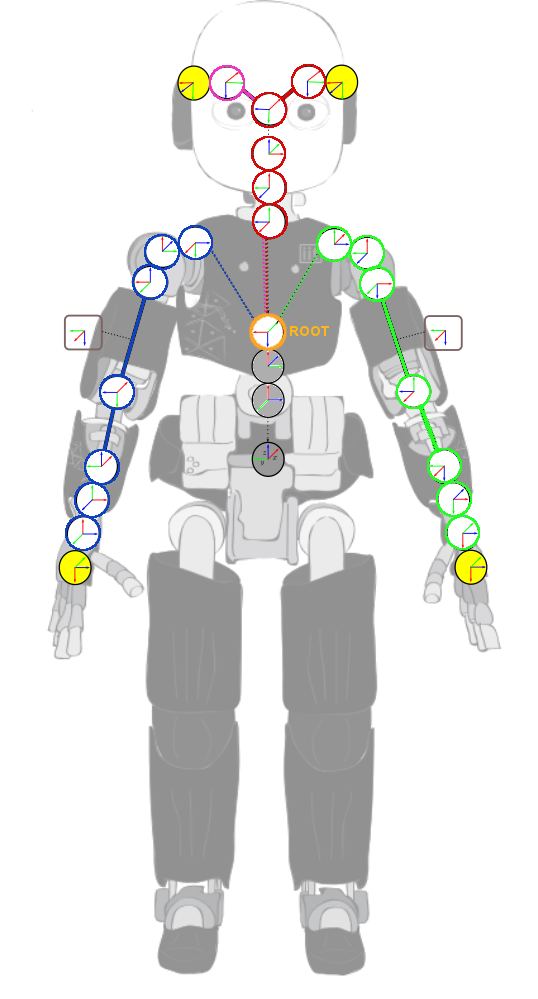} \includegraphics[width = 125 pt]{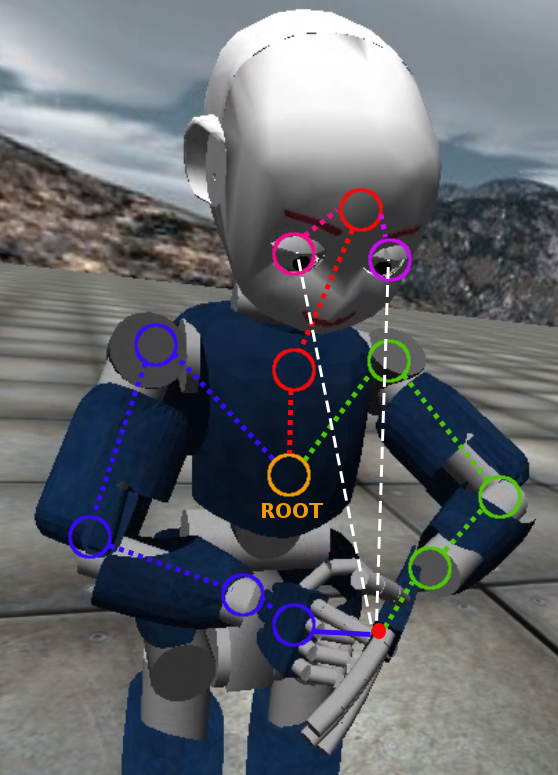}}}
      \caption{iCub upper body and schematic illustration of kinematic chains considered. All chains originate in a common \textit{Root} which is located at the third torso joint. The left and right arm chains are  drawn in green and blue respectively. The eye chains have a common Root-to-head chain part marked in red. The right panel illustrates the self-calibration by connecting different chains---self-touch and self-observation. White lines denote projection into the eyes/cameras.}
\label{fig:kinModel}
\end{figure}

In this work, we use the upper body of the iCub humanoid robot (see Fig. \ref{fig:kinModel}) and its kinematic model expressed in the Denavit-Hartenberg convention, where every link $i$ is described by 4 parameters: $\{a_i, d_i, \alpha_i, o_i\}$. In this platform, all joints are revolute. We will consider several kinematic chains: all start in a single inertial or base frame---denoted iCub \textit{Root} Reference Frame here.
For every chain, the DH parameters uniquely define a chain of transformation matrices from the inertial frame to the end-effector. The position and orientation of the end-effector in the \textit{Root} frame is thus given by ${\boldsymbol{T}}_n^{Root} = A_1(q_1)...A_n(q_n)$ where the homogeneous transformation matrices $A_i$ can be constructed from the DH representation and $q_i$ are current joint angles of the robot actuators. 

The links are schematically illustrated in Fig. \ref{fig:kinModel}. iCub kinematics version 1 was used \cite{icubWiki} with the following modification: the \textit{Root} was moved from the waist area to the third torso joint, which is the new inertial frame for our purposes. 

The four chains under consideration are:
\begin{enumerate}
\item Left Arm (LA). DH parameters in Table~\ref{tab:LA_DH}. Short names to denote the links/joints: Root-to-LAshoulder, LA Shoulder Pitch, LA Shoulder Roll, LA Shoulder Yaw, LA Elbow, LA Wrist Prosup (for pronosupination), LA Wrist Pitch, LA Wrist Yaw.
\item Right Arm (RA). DH parameters analogous to LA (see~\cite{icubWiki}). Link/joint names: Root-to-RAshoulder, RA Shoulder Pitch, RA Shoulder Roll, RA Shoulder Yaw, RA Elbow, RA Wrist Prosup, RA Wrist Pitch, RA Wrist Yaw.
\item Left Eye (LEye). DH parameters in Table~\ref{tab:LEye_DH}. Link/joint names: Root-to-neck, Neck Pitch, Neck Roll, Neck Yaw, Eyes Tilt, Left Eye Pan.
\item Right Eye (REye). DH parameters different than LEye in Table~\ref{tab:REye_DH}. Link/joint names: Root-to-neck, Neck Pitch, Neck Roll, Neck Yaw, Eyes Tilt, Right Eye Pan.
\end{enumerate}

Links or parameters not subject to calibration are showed shaded in grey in the corresponding tables. The first link always originates in the Root frame and is fixed in all chains (the torso joint is not moving) and is also excluded from calibration. The alpha parameter of the last link in the arm chains is also not being calibrated as it is not observable because we observe only position and not the orientation of the end-effectors. The right arm chain is further extended with a fixed transform from the end-effector in the palm to the tip of the index finger---not subject to calibration. The eye chains differ in the last link only. 

\begin{table}[htpb]
\centering
\begin{tabular}{c|cccc}
\hline
Link(i) & a(i) [mm] & d(i) [mm] &$\alpha$ [rad] & $o$ [rad]\\
\hline
\rowcolor{Gray}
1 & 23.36 & 143.3 & $\pi/2$ & $105*\pi/180°$\\
2 & 0 & 107.74 & $-\pi/2$ & $\pi/2$\\
3 & 0 & 0 & $\pi/2$ & $-\pi/2$\\
4 & 15 & 152.28 & $-\pi/2$ & $75*\pi/180°$\\
5 & -15 & 0 & $\pi/2$ & 0\\
6 & 0 & 137.3 & $\pi/2$ & $-\pi/2$\\
7 & 0 & 0 & $\pi/2$ & $\pi/2$\\
8 & 62.5 & -16 & \cellcolor{Gray}0 & 0\\
\end{tabular}
\caption{DH parameters ($a, d, \alpha$ and offsets $o$) describing all links in Left Arm kinematic chain.}
\label{tab:LA_DH}
\end{table}

\begin{table}[htpb]
\centering
\begin{tabular}{c|cccc}
\hline
Link(i) & a(i) [mm] & d(i) [mm] &$\alpha$ [rad] & $o$ [rad]\\
\hline
\rowcolor{Gray}
1 & 2.31 & - 193.3 & $-\pi/2$ & $\pi/4$\\
2 & 33 & 0 & $\pi/2$ & $\pi/4$\\
3 & 0 & 1 & $-\pi/2$ & $\pi/4$\\
4 &- 54 & 82.5 & $-\pi/2$ & $\pi/4$\\
5 & 0 & - 34 & $-\pi/2$ & $0$\\
6 & 0 & 0 & $\pi/2$ & $-\pi/4$\\
\end{tabular}
\caption{DH parameters -- Left Eye kinematic chain.}
\label{tab:LEye_DH}
\end{table}

\begin{table}[htpb]
\centering
\begin{tabular}{c|cccc}
\hline
Link(i) & a(i) [mm] & d(i) [mm] &$\alpha$ [rad] & $o$ [rad]\\
\hline
5 & 0 &  34 & $\pi/2$ & $-\pi/4$\\
6 & 0 & 0 & $-\pi/2$ & $0$\\
\end{tabular}
\caption{DH parameters -- Right Eye kinematic chain. Links 1-4 shared with Left Eye kinematic chain.}
\label{tab:REye_DH}
\end{table}

The camera intrinsic parameters were taken from the real robot cameras and were not subject to calibration: resolution $320$ x $240$, focal length $f_x=257.34$, $f_y=257.34$.%, optical center $c_x=160$, 
$c_y=120$.

\subsection{Optimization problem formulation}
\label{sec:optim}

By calibration we mean estimation of the parameter vector ${\boldsymbol{\phi}} =\{
[a_1,...,a_n], [d_1,...,d_n], [\alpha_1,...,\alpha_n], [o_1,...,o_n]\}$ with $i \in N$, where $N = \{1,..,n \}$ is a set of indices identifying individual links; $a$, $d$ and $\alpha$ are the first three parameters of the DH formulation and $o$ the offset that specifies the positioning of the encoders on the joints with respect to the DH representation. We often estimate a subset of these parameters only,  assuming that the others are known. This subset can for example consist of a subset of links $N' \subset N$ (e.g., only parameters of one arm are to be calibrated) or a subset of the parameters (e.g., only offsets $o$ are to be calibrated---sometimes dubbed ``daily calibration'' \cite{Nickels2003}).

The estimation of the parameter vector $\boldsymbol{\phi}$ is done by optimizing a given objective function: 
\begin{equation}
\label{eq:optim}
{\boldsymbol{\phi}}^* = \argmin_{\boldsymbol{\phi}} \sum_{m=1}^M || {\boldsymbol{p}}_m^r - {\boldsymbol{p}}_m^e ({\boldsymbol{\phi}}, {\boldsymbol{\Theta}}_m)||^2,
\end{equation}
where $M$ is the number of robot configurations and corresponding end-effector positions used for calibration (hereafter, often referred to as ``poses'' for short), ${\boldsymbol{p}}_m^r$ is a real (observed) end-effector position, ${\boldsymbol{p}}_m^e$ is an estimated end-effector position computed using forward kinematic function for a given parameter estimate ${\boldsymbol{\phi}}$ and joint angles from joint encoders ${\boldsymbol{\Theta}}_m$.
For chains involving cameras, the reprojection error is used instead, as described in the next section.

\subsection{Kinematic chain calibration}
We study different combinations of intersecting chains and their performance in calibrating one another.
\subsubsection{Two arms chain (LA-RA)}
This corresponds to the self-touch scenario, with touch occurring directly at the end-effectors (the right arm end-effector being shifted from palm to tip of index finger using a fixed transform). The newly established kinematic chain for upper body includes both arms while head and eyes are excluded. To optimize parameters describing this chain, we minimize the distance between estimated positions in 3D space of left and right arm end-effectors. In this case, the parameter vector ${\boldsymbol{\phi}}$ consists of the following parameters: ${\boldsymbol{\phi}} = \{{\boldsymbol{\phi}}^r,{\boldsymbol{\phi}}^l\}$, where ${\boldsymbol{\phi}}^r$ and $\boldsymbol{\phi}^l$ are parameters corresponding to the robot right and left arm, respectively. The objective function to be optimized is:% then defined as follows:
\begin{equation}
\label{eq:lara}
{\boldsymbol{\phi}}^* = \argmin_{\boldsymbol{\phi}}\sum_{m=1}^M||{\boldsymbol{X}}_{m}^{r,R} ({\boldsymbol{\phi}}^r, {\boldsymbol{\Theta}}_m^r) - {\boldsymbol{X}}_{m}^{l,R} ({\boldsymbol{\phi}}^l, {\boldsymbol{\Theta}}_m^l)||^2
\end{equation}
where $M$ is the number of poses used for calibration, ${\boldsymbol{X}}_{m}^{r,R}$ and ${\boldsymbol{X}}_{m}^{l,R}$ are the $m$\textsuperscript{th} estimated end-effector positions in the Root frame for the right and left arm respectively, computed using a given parameter estimate $\boldsymbol{\phi}$ and joint angles from joint encoders ${\boldsymbol{\Theta}}_m$.

\subsubsection{Hand to eye chains (LA-LEye, LA-REye, RA-LEye, RA-REye)}

To predict position of the end-effector in each of the robot cameras (similar to \cite{Birbach2015}), the estimated end-effector position, ${\boldsymbol{X}}^{Root}$, is given by a current hypothetical robot calibration of the parameter vector ${\boldsymbol{\phi}}$ and is computed via forward kinematics. ${\boldsymbol{X}}^{Root}$ is then mapped to left camera coordinates (${\boldsymbol{X}}^{LEye}$) using a transformation matrix ${\boldsymbol{T}}_{Root}^{LEye}$. Then we use a pinhole camera model to transform the 3D point (${\boldsymbol{X}}^{LEye}$) into image coordinates (${\boldsymbol{X}}^{img}$):
\begin{equation}
\begin{pmatrix}
X^{img}_x \\ 
X^{img}_y
\end{pmatrix}
 = 
\begin{pmatrix}
f_x X^{LEye}_x/X^{LEye}_z \\ 
f_y X^{LEye}_y/X^{LEye}_z 
\end{pmatrix},
\end{equation}
where $f_x$, $f_y$ are focal lengths of the camera. Radial distortion of cameras was not considered. 

This approach doesn't require information from both eyes and enables us to estimate only one side of the robot body (e.g. parameters of the left arm and left eye). For example, the estimated parameter vector $\boldsymbol{\phi}$ in the case of the kinematic chain connecting left arm and left eye consists of the following parameters: ${\boldsymbol{\phi}} = \{{\boldsymbol{\phi}}^l,{\boldsymbol{\phi}}^{le}\}$, where  ${\boldsymbol{\phi}}^l$ and ${\boldsymbol{\phi}}^{le}$ are parameters corresponding to the robot left arm and to the left eye, respectively. The objective function is then defined as:
\begin{equation}
\label{eq:laley}
{\boldsymbol{\phi}}^* = \argmin_{\boldsymbol{\phi}} \sum_{m=1}^M || {\boldsymbol{X}}^{l,img}_m({\boldsymbol{\phi}}^l,{\boldsymbol{\phi}}^{le}) - {\boldsymbol{u}}^L_m||^2,
\end{equation}
where ${\boldsymbol{X}}^{l,img}_m$ is the $m$\textsuperscript{th} 2D position of the estimated left arm end-effector projected to left eye image coordinates and ${\boldsymbol{u}}^L_m$ is the $m$th 2D position of the observed left arm end-effector in the left camera. For two arms and two eyes we get four possible combined chains: left/right arm to right/left eye. Since the results are similar due to symmetry, we present in the experimental section results only for the Left arm - Left eye (LA-LEye) chain.

\subsubsection{Combining multiple chains (LA-RA-LEye, LA-RA-LEye-REye)}
In order to estimate all kinematic parameters of the robot, we can take advantage of combining some or all of the above mentioned kinematic chains. 
For example, in the case that we combine LA-RA, LA-LEye and LA-REye chains together into LA-RA-LReye, the estimated parameter vector $\boldsymbol{\phi}$ consists of the following parameters: ${\boldsymbol{\phi}} = \{{\boldsymbol{\phi}}^r,{\boldsymbol{\phi}}^l,{\boldsymbol{\phi}}^{re},{\boldsymbol{\phi}}^{le}\}$, where ${\boldsymbol{\phi}}^l$, ${\boldsymbol{\phi}}^r$, ${\boldsymbol{\phi}}^{re}$,  and ${\boldsymbol{\phi}}^{le}$ are parameters corresponding to the left arm, right arm, right eye, and left eye, respectively. The objective function is in this case defined as:

\begin{equation}
\label{eq:laralreye}
\begin{split}
{\boldsymbol{\phi}}^* =& \argmin_{\boldsymbol{\phi}} \sum_{m=1}^M \{\mu\cdot||{ {\boldsymbol{X}}}_{m}^{r,R} ({\boldsymbol{\phi}}^r, {\boldsymbol{\Theta}}_m^r) - {\boldsymbol{X}}_{m}^{l,R} ({\boldsymbol{\phi}}^l, {\boldsymbol{\Theta}}_m^l)||+\\
& || {\boldsymbol{X}}^{l_L,I}_m({\boldsymbol{\phi}}^l,{\boldsymbol{\phi}}^{le}) - {\boldsymbol{u}}^{l_L}_m|| + || {\boldsymbol{X}}^{r_L,I}_m({\boldsymbol{\phi}}^r,{\boldsymbol{\phi}}^{le}) - {\boldsymbol{u}}^{r_L}_m|| +\\
& || {\boldsymbol{X}}^{l_R,I}_m({\boldsymbol{\phi}}^l,{\boldsymbol{\phi}}^{re}) - {\boldsymbol{u}}^{l_R}_m|| + || {\boldsymbol{X}}^{r_R,I}_m({\boldsymbol{\phi}}^r,{\boldsymbol{\phi}}^{re}) - {\boldsymbol{u}}^{r_R}_m||\}^2,\\
\end{split}
\end{equation}
where $M$ is the number of poses (configurations) used for calibration, ${\boldsymbol{X}}_{m}^{r,R}$ and ${\boldsymbol{X}}_{m}^{l, R}$ are the $m$\textsuperscript{th} estimated end-effector positions in the Root frame for the right and left arm, respectively. These are computed using a given parameter estimate $\boldsymbol{\phi}$ and joint angles from joint encoders ${\boldsymbol{\Theta}}_m$. Values ${\boldsymbol{X}}^{l_L,I}_m$ and ${\boldsymbol{X}}^{r_L,I}_m$  are the $m$\textsuperscript{th} positions of the estimated left arm  end-effector projected to left eye and right eye image coordinates, respectively, and ${\boldsymbol{u}}^{l_L}_m$ and ${\boldsymbol{u}}^{r_L}_m$ are the $m$\textsuperscript{th} 2D position (pixel coordinates) of the left arm end-effector observed in the left and right eye/camera, respectively (variables ${\boldsymbol{X}}^{l_R,I}_m$, ${\boldsymbol{X}}^{r_R,I}_m$, ${\boldsymbol{u}}^{l_R}_m$ and ${\boldsymbol{u}}^{r_R}_m$ correspond to the right arm). Since the cost function contains both 3D and reprojection errors, the distances in space were multiplied by a coefficient $\mu$ determined from the intrinsic parameters of cameras and distance $d$ of the end-effector from the eye: $\mu = 320px/(d* (\pi/3)) $.

\subsection{Non-linear least squares optimization}
The objective functions  (Eqs.~[\ref{eq:optim}]- [5]) defined for the optimization problem described in Section~\ref{sec:optim} are of the least-squares form and therefore can be minimized by Levenberg-Marquardt algorithm for non-linear least squares optimization (we used MATLAB implementation of the algorithm, same as in \cite{Birbach2015}). This iterative local algorithm performs minimization of a non-linear objective function by linearizing it at the current estimate every iteration. It interpolates between the Gauss-Newton and gradient descent method, combining advantages of both.

\subsection{Error metrics}
For comparing the results achieved for individual settings, we make use of the following error metrics:
\subsubsection{Cartesian error between poses (position)}
Cartesian position error $E_{c}$ between two generic poses, A and B, where ${\boldsymbol{P_A}} = [x_A, y_A, z_A]$ and ${\boldsymbol{P_B}} = [x_B, y_B, z_B]$ are 3D Cartesian positions of the end-effector, is defined as:
\begin{equation}
\begin{split}
E_c = \sqrt{(x_A-x_B)^2+(y_A-y_B)^2+(z_A-z_B)^2}.
\end{split}
\end{equation}
We evaluate the Cartesian error over the set of $N$ testing poses, which are selected as described in the section~\ref{sec:ttdata}.

\subsubsection{Quality of estimated parameters} For each estimated parameter $\phi_i$ we compute the mean difference ($e_i$)  of the estimated parameter $\phi_i^e$ from the target parameter value $\phi_i^t$  (averaged over $R$ repetitions of the experiment): 
\begin{equation}
e_i = {{\sum_{r=1}^R{ |\phi_{i,r}^e-\phi_i^t|}}\over{R}},
\end{equation}
as well as standard deviation of the parameter.

\section{Data acquisition and description}
\subsection{Pose set generation}
\label{subsec:pose_set_gen}
With the goal of comparing different calibration methods on a humanoid robot, we chose a dataset where the two arms of the robot are in contact---thereby physically closing the kinematic chain through self-touch. At the same time, the robot gazes at the contact point (self-observation). 
The points were chosen from a cubic volume in front of the robot. For each target, using the Cartesian solver and controller \cite{Pattacini2010}, the iCub moves the left hand with end-effector in the palm to the specified point. Then it moves the right hand, with end-effector in the tip of the index finger, to the same point, with the additional constraint that the finger can be at most $50^\circ$ away from the perpendicular direction of the palm. 5055 points and corresponding joint configurations were thus generated, with a difference on left and right effector position in every configuration of maximum $0.01 mm$---see Fig.~\ref{fig:poses_vis}, right. The gaze controller \cite{Roncone2016gaze} was used to command the neck and eyes of the robot to gaze at the same target (code and video can be accessed at \cite{github-datasetGenerator}). The full dataset thus consists of 5055 data vectors ${\boldsymbol{X}}_i = [{\boldsymbol{X}}^{target}_i, {\boldsymbol{X}}^{RA}_i, {\boldsymbol{X}}^{LA}_i,
{\boldsymbol{\Theta}}_i]$ composed of target point coordinates (${\boldsymbol{X}}^{target}_i \in \mathbb{R}^3$), corresponding right arm and left arm end-effector positions (${\boldsymbol{X}}^{RA} \in \mathbb{R}^3$, ${\boldsymbol{X}}^{LA} \in \mathbb{R}^3$), and joint angles $\boldsymbol{\Theta}_i$ for every joint of the torso, arms, neck, and eyes (${\boldsymbol{\Theta}}_i \in \mathbb{R}^{20}$). 
Note that the solvers work with a given tolerance and hence 
$ X^{target}_i \neq X^{RA}_i \neq X^{LA}_i$. 

This way of dataset generation draws on previous work \cite{Roncone_ICRA_2014} and is hence feasible on the real robot provided sufficient quality of the initial model. Li et al. \cite{QiangLi2015} provide an alternative control method: ``tactile servoing''. The robot could be also manipulated into the desired configurations while in gravity compensation mode.

\subsection{Training and testing dataset}
\label{sec:ttdata}
\begin{figure}[thpb]
    \centering
    \framebox{\parbox{3.3in}{\includegraphics[width=240 pt]{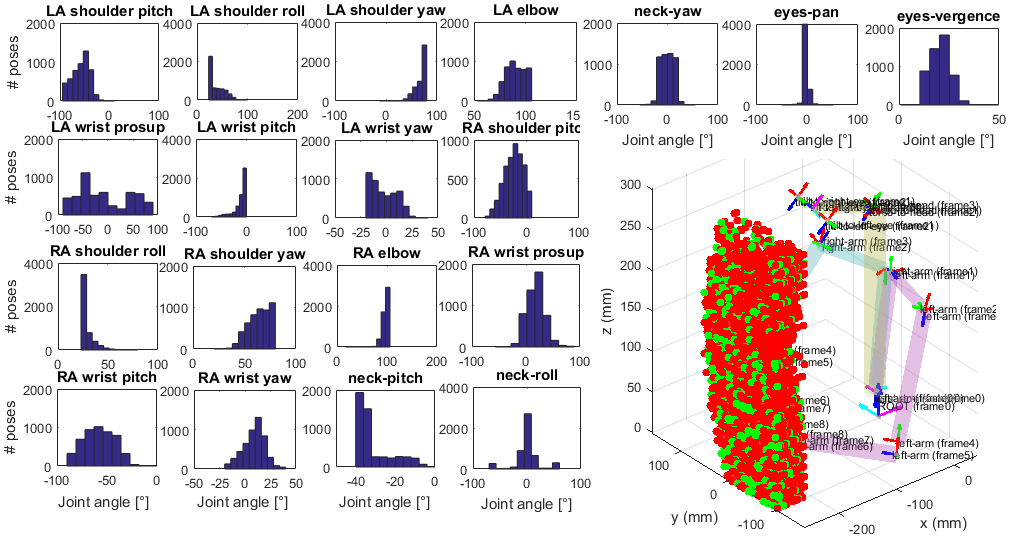}}}
      \caption{Dataset visualization -- 5055 configurations. (left) Distribution of joint values. (right) End-effector positions. Red -- left arm; Green -- right arm. 
      }
     \label{fig:poses_vis}
\end{figure}

We had 5055 configurations with $|\boldsymbol{X}^{RA}_i - \boldsymbol{X}^{LA}_i| < 0.01$ $mm$. The $0.01$ $mm$  error will at the same time constitute the lower bound on the maximum achievable calibration accuracy using the closure of the kinematic chain through self-touch. For the case of loop closure through the cameras, we employ the neck and eye joint values obtained from the solver in the simulator but reproject the end-effector positions directly and accurately into the cameras simulated in Matlab. The 5055 data points were further divided into training and testing datasets in the following way: $N$ out of 4755 poses are used as a training set on which the optimization process is performed (with a subset of 10, 20, 50, or 1000 poses chosen at random in different experiments) and 300 poses are used for testing purposes. Fig. \ref{fig:poses_vis}, left, shows the distribution of joint values for individual joints in the dataset---this may impact the identifiability of individual parameters.

%\vspace{-0.5cm}
\subsection{Measurement error}
\label{sec:MeasurementError}
Measurement noise with a Gaussian distribution was added motivated by the sensory accuracy in the real robot. Since distance between individual taxels on the real iCub sensitive skin is around 5 mm, we decided to use Gaussian noise with zero mean and $\sigma^2 = 5$ for touch as a baseline. For cameras, we introduce a 5px error (Gaussian noise with zero mean and $\sigma^2 = 5$ px), inspired by the setup in \cite{Fanello2014} where the iCub is detecting its fingertip in the camera frame. These errors are used in all experiments in the Simulation results section if not stated otherwise. In Section~\ref{sec:measError} we evaluate how  changing the size of these measurement errors affects the resulting accuracy of end-effector position detection for individual chains.

\subsection{Perturbation of the initial parameters estimate}
To evaluate the dependence of the optimization performance on the quality of the initial estimates of the parameters, we perturbed all estimated parameters by a \textit{perturbation factor} $p = \{2,5,10,20\}$. We perturbed all initial offset values $o_i$ as follows:
\begin{equation}
o^{new}_i = 1/100*p*uniform[-1;1]+ o_i \: [rad],
\end{equation}
It is reasonable to expect that the remaining DH parameters ($\alpha$, $a$, and $d$) will be in general more accurate as they can be extracted from CAD models and there is no moving part and no encoder involved. Therefore, their perturbation was chosen as follows:
\begin{equation}
\begin{split}
&\alpha: \alpha^{new}_i = 1/1000*p*uniform[-1;1]+\alpha_i \: [rad],\\
&a, d: \Phi^{new}_i = 0.1*p*uniform[-1;1]+\Phi_i \: [mm].\\
\end{split}
\end{equation}

\section{SIMULATION RESULTS}
In this section we show the calibration results. We evaluated our approach using both error of the end-effector position---the cost function optimized (or distance in camera frame for projections into eyes)---as well as error in individual parameters (vs. their correct values). We compared kinematic chains used for calibration, number of free parameters which were estimated by the optimization process, different perturbation factor on individual parameters, number of training poses (data points), as well as measurement noise levels. Performance is always is evaluated on the testing dataset. 

\subsection{Results for different chain combinations and number of training poses}
\begin{figure}[htb]
    \centering
    \includegraphics[width=240 pt]{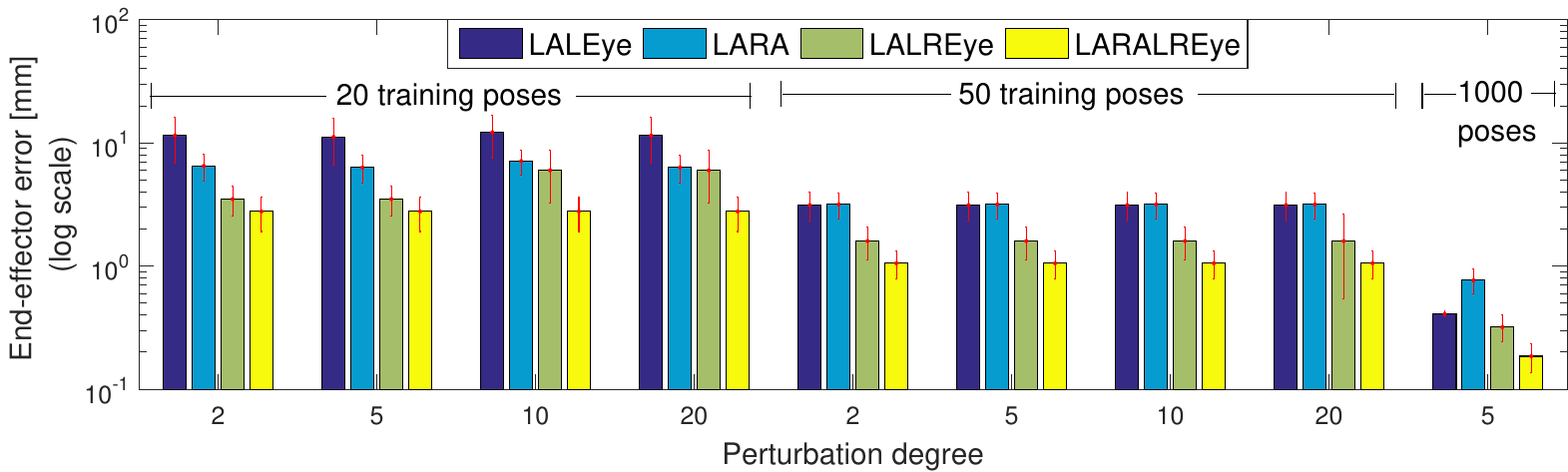}
         \includegraphics[width = 240pt]{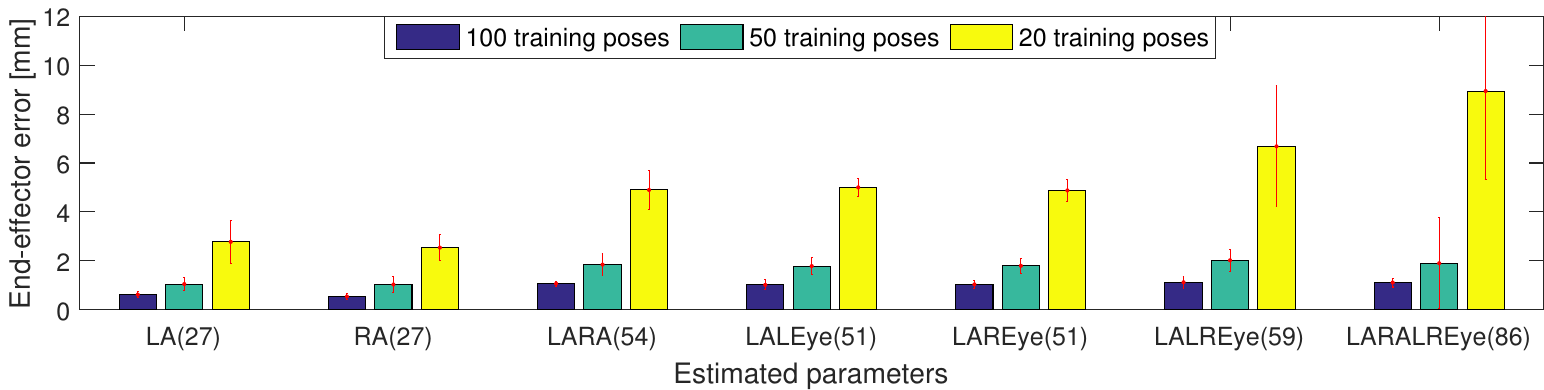}
    \caption{End-effector position error after optimization---averaged over 10 repetitions. (Top) Left Arm chain calibration (full DH) using different chain combinations,  different initial perturbation factors (2, 5, 10, 20)  and training on 20 (left), 50 (middle), and 1000 poses (right -- pert. factor 5 only).  (Bottom) Performance of different parameter sets subject to calibration -- LARALREye chain was used for calibration of parameters. Free parameters (being calibrated) in a given chain are denoted. E.g., LALEye denotes that all 51 DH parameters of left arm and left eye (including head) are calibrated, and the rest of the DH parameters (e.g. right arm) is considered to be known.}
    \label{fig:pertDeg}
\end{figure}

Fig.~\ref{fig:pertDeg} (top) shows the performance in terms of end-effector position estimation when DH parameters of the left arm (LA) chain are calibrated, utilizing different kinematic chain combinations: The ``self-observation'' from a single camera (LALEye) and ``self-touch'' only (LARA) are outperformed by ``stereo self-observation'' (LALREye) and all the chains together provide the best results (LARALREye). Clearly, more training poses (50 vs. 20) improve calibration results; 1000 poses should be sufficient to reach an optimal value and serve as a lower bound on the error. The effect of initial parameter perturbation factor is also shown;  for all perturbation levels, the performance is stable (low error variance). 

In Fig.~\ref{fig:pertDeg} (bottom) only the largest ``multi-chain'' LARALREye is employed for training but the chains whose parameters are subject to calibration are varied. The error of end-effector position estimation is increasing with higher number of parameters estimated; however, even if parameters of all chains (86 DH parameters) are perturbed and subject to calibration simultaneously, end-effector error of around 2 (1) $mm$ can be achieved with 50 (100) poses.
 
\begin{figure}[thpb]
    \centering
      \framebox{\parbox{3.3in}{\includegraphics[width=240 pt]{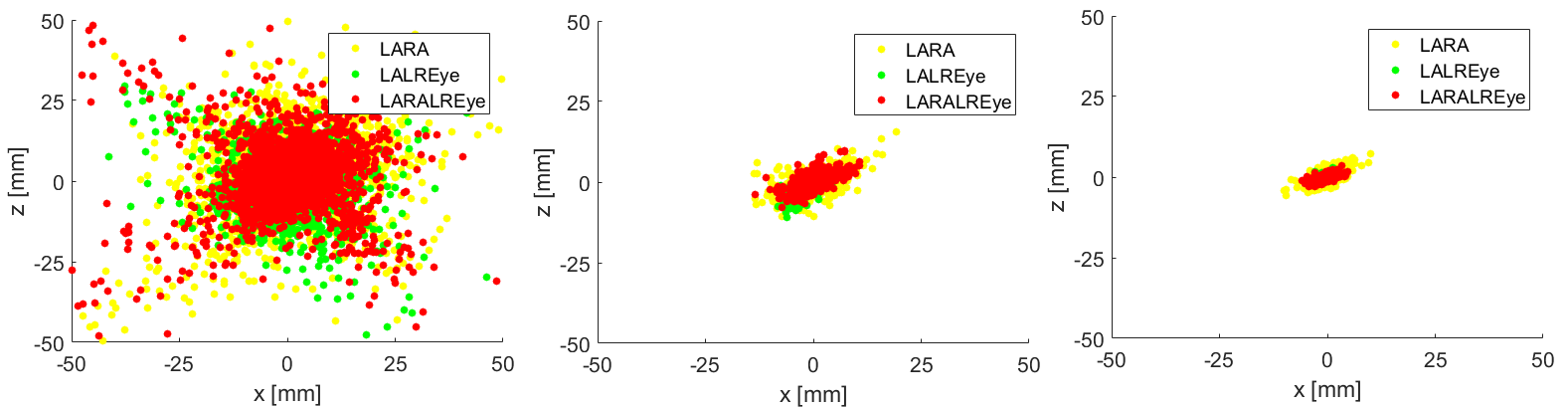}}}
      \caption{Error residuals -- Left Arm (LA) chain calibration using LARA, LALREye and LARALREye chains. Results visualized on 300 testing poses for each of 10 repetitions of the optimization, with random parameter initialization (3000 points in total per chain shown). (Left) 10 training poses; (Middle) 20 training poses; (Right) 50 training poses. Perturbation factor 10 and measurement errors 5 mm for skin and 5 px for cameras were considered.}
      \label{fig:chains}
\end{figure}

To investigate the distribution of errors for individual chains, we examined error residuals for every testing pose. For a higher number of training poses, error residuals have a zero mean and Gaussian distribution. For lower number of poses (especially for higher perturbation), the residuals are bigger and skewed and the resulting calibration also strongly depends on initialization. In Fig.~\ref{fig:chains}, the end-effector error residuals for perturbation factor $p=10$ are shown 
for their $x$ and $z$ coordinates (other 2D projections were qualitatively similar)---for different chains and different number of training poses. 

\subsection{Observability analysis of individual chains}
\begin{figure}[thpb]
    \centering
      \includegraphics[width=230pt]{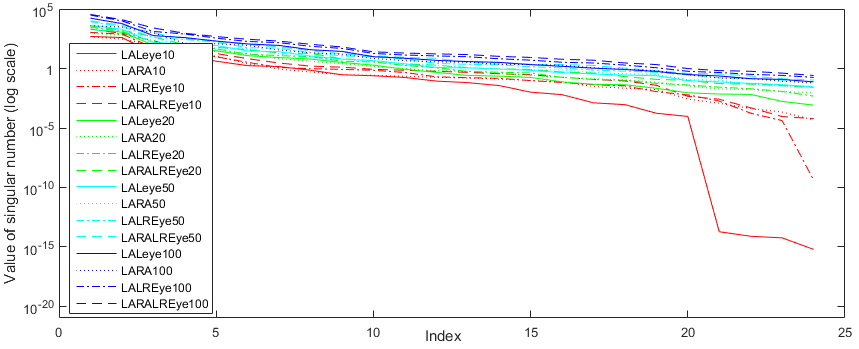}
      \includegraphics[width=120pt]{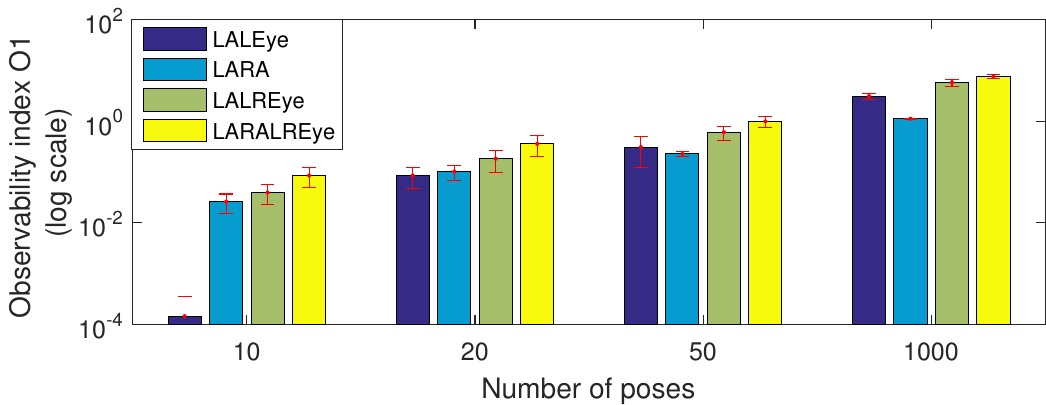}
      \includegraphics[width=120pt]{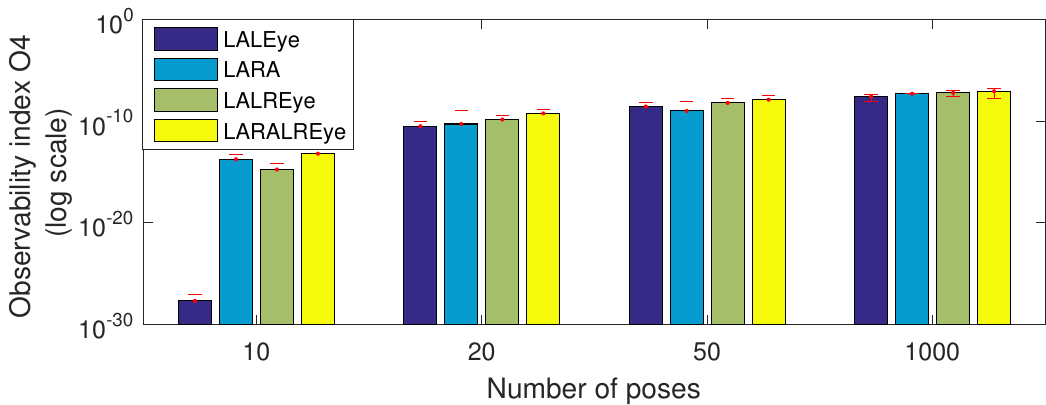}
      \caption{Observability -- Left Arm (LA) chain calibration (full DH) using different chain combinations. (Top) singular numbers of identification Jacobian for different chains used for calibration. Evaluation is performed over the same pose set for every chain. Red, green, turquoise, and blue color of the lines denote 10, 20, 50, and 1000 poses in the training set respectively. (Bottom left) Observability index $O_1$~\cite{borm1989}. (Bottom right) Observability index $O_4$~\cite{nahvi1996}.}
      \label{fig:observability}
\end{figure}

We conducted an observability analysis using Singular Value Decomposition (SVD) of the identification Jacobian matrix $J = [J_1,...,J_n]$, where $n$ is the number of configurations in the training pose set and $J_n(i,j)=\left[\partial (X^r_i-X^e_i) \over \partial{\phi_j}\right]$, $\phi_j$ is the parameter $j$ to be estimated, $(X^r_i-X^e_i)$ denotes the error between the real/observed ($X^r$) and estimated ($X^e$) value of the $i$\textsuperscript{th} coordinate in the given chain.\footnote{ E.g., for LALEye, $X$ corresponds to 2 errors: error on the coordinate $u$ and $v$ as a reprojection of the end-effector position into the cameras; for LARA chain, $X$ will correspond to 3 numbers: distance in x, y and z coordinate between right ($X^{r,R}$) and left arm ($X^{l,R}$) end-effector 3D positions.} The Jacobian matrix represents the sensitivity of end-effector positions or their camera reprojections to the change of individual parameters. Using SVD, we can obtain a vector of singular numbers $\sigma_i$. Comparison of the obtained singular numbers for individual chains for the task of estimating all DH parameters of the left arm (using same training pose set) can be seen in Fig.~\ref{fig:observability}. 
We also evaluated observability indices $O_1$~\cite{borm1989} and $O_4$~\cite{nahvi1996} (performance of observability indices for industrial robot calibration was evaluated by Joubair~\cite{joubair2016}). $O_1$ index is defined as: $O_1 = {(\sigma_1 \sigma_2...\sigma_m)^{1/m} \over {\sqrt(n)}}$, where $m$ is the number of independent parameters to be identified, $\sigma_i$ is the $i$\textsuperscript{th}  singular number, and $n$ is the number of calibration configurations. Index $O_4$ is defined as: $\sigma_m^2 \over \sigma_1$. See Fig.~\ref{fig:observability} (bottom panels). 
Te chain LALEye for 10 poses has very low observability caused by not full rank Jacobian (we have 24 parameters to estimate but only 20 equations). The highest observability is achieved in all cases for the largest chain LARALREye, where the information from touch and both cameras was used.

\subsection{Evaluation of error based on measurement noise}
\label{sec:measError}
\begin{figure}[thpb]      
   \centering
  \framebox{\parbox{3.2in}{\includegraphics[width=230 pt]{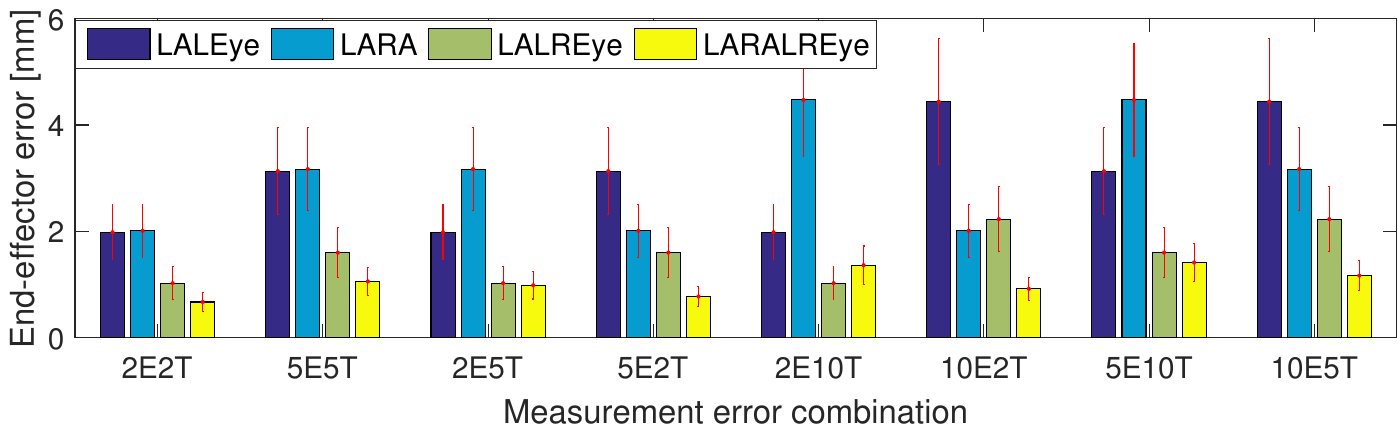}}}
      \caption{End-effector position accuracy for different combinations of measurement noise on cameras and ``touch sensor''. Different chains employed to estimate DH parameters of the left arm (50 training poses, error evaluated over 300 testing poses, averaged over 10 repetitions). X-axis labels read as follows: first number -- error on cameras (``Eyes'') in pixels;   second number -- error on the touch sensor in $mm$ (i.e. 5E2T denotes that we introduced zero-mean Gaussian error with 5px and 2mm variance to cameras and touch respectively. 
      }     
       \label{fig:MeasurementError}
   \end{figure}

We evaluated the effect of measurement noise in individual sensors (touch, cameras) on the accuracy of end-effector position error on the testing data set---see Fig.~\ref{fig:MeasurementError}. With same error in pixels on cameras and in $mm$ on ``touch sensors'' (first two columns -- $2px$/$2mm$, $5px$/$5mm$), LALREye chain (both eyes, no touch) and LARALREye (both eyes and touch) have smallest final end-effector errors, for the ``multi-chain'' even smaller. When error on cameras increases ($5E2T$, $10E2T$, $10E5T$), the camera chains (LALEye, LALREye) are affected whereas the performance of the chain with touch (LARALREye) is not degraded. Conversely, more error on ``touch'' ($2E5T$, $2E10T$, $5E10T$) impacts the ``touch only'' chain (LARA), but the LARALREye remains robust. 
 
\subsection{Quality of DH parameter estimates}
To get further insight and take advantage of the simulation study where we have access to ground truth values of all parameters, we also studied whether the optimization based on end-effector error also leads to correct estimates of all DH parameters---focusing on the left arm (LA) chain. 

\begin{figure}[thpb]      
       \centering
      \includegraphics[width=230 pt]{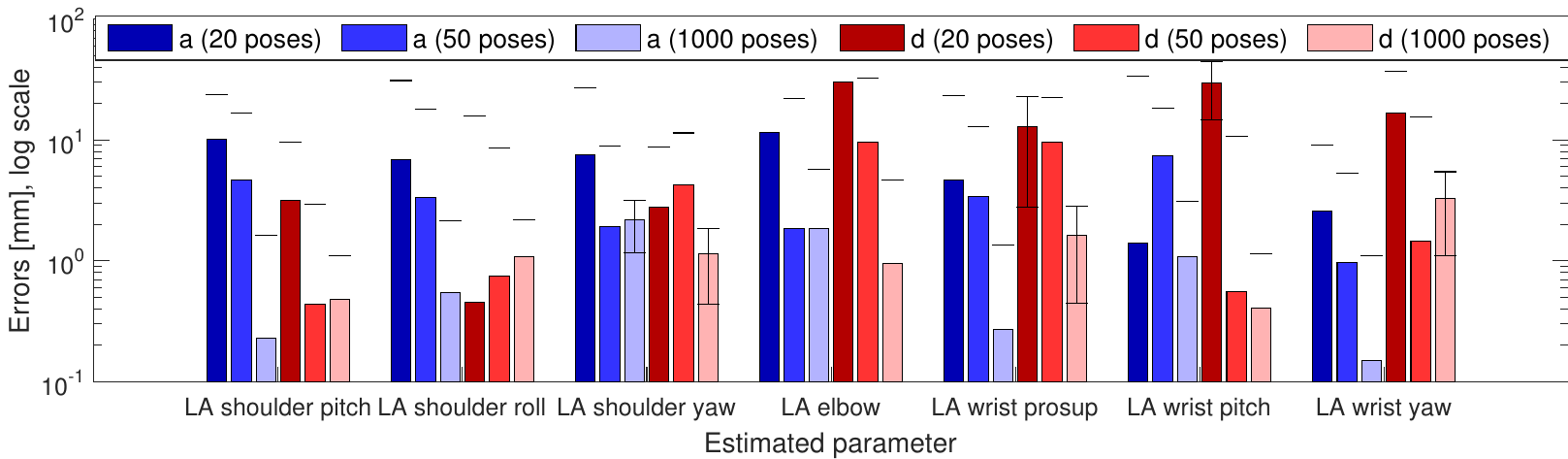}
      
      \includegraphics[width=230 pt]{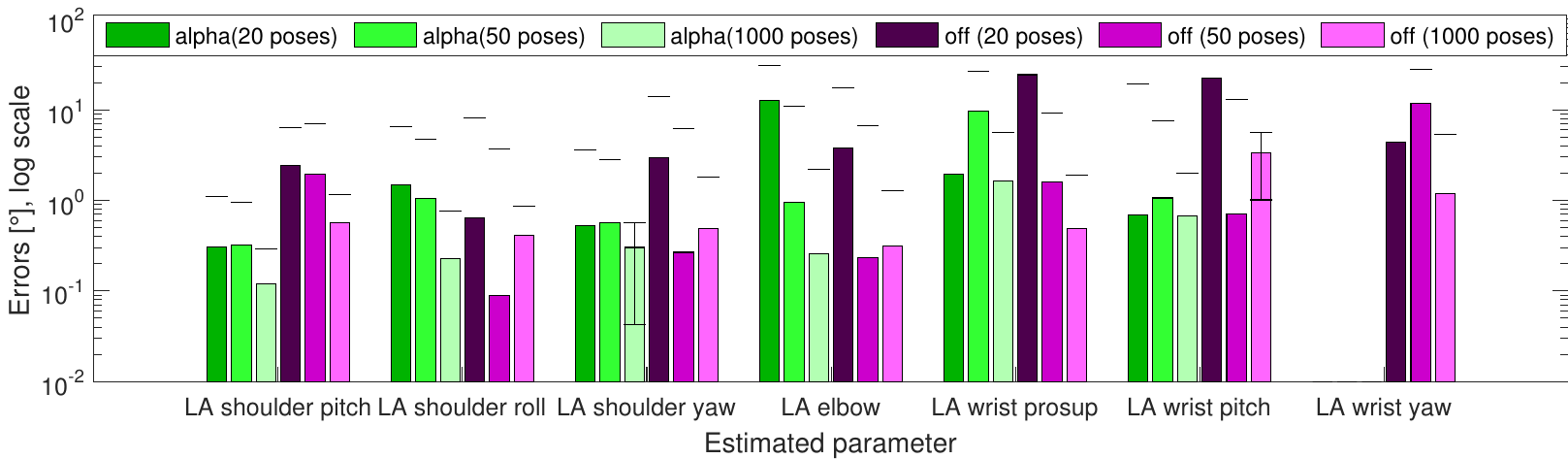}
      \caption{Quality of DH parameter estimation for LA chain using LA-RA chain. Errors on individual parameters after optimization for different number of poses: (Top) $a$ and $d$  parameters; (Bottom) $\alpha$ and \textit{offsets}. Averaged over 10 repetitions, perturbation factor 5, measurement noise $5px$ on cameras and $5mm$ on touch.}     
       \label{fig:parsLARA}
   \end{figure}

Fig.~\ref{fig:parsLARA} shows the results for all estimated parameters when the LA-RA (``self-touch'') chain was used for calibration, using different number of training poses. The errors on the length parameters (top panel) are on average distributed between approx. 1 and 10 $mm$. For the angular quantities, it is in the $0.1$ to $1 ^\circ$ range for the proximal joints.

\begin{figure}
    \centering
    \includegraphics[width=240 pt]{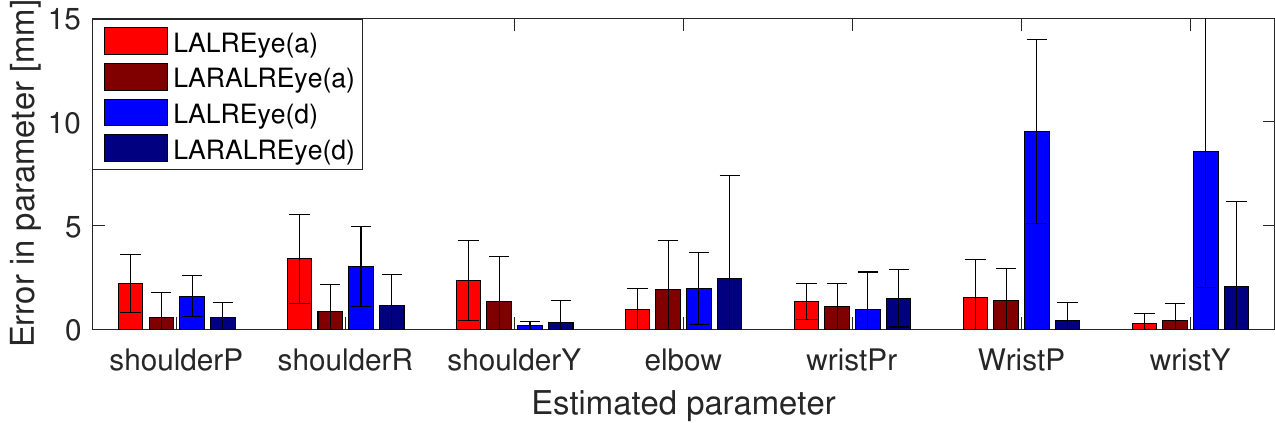}
    \includegraphics[width=240 pt]{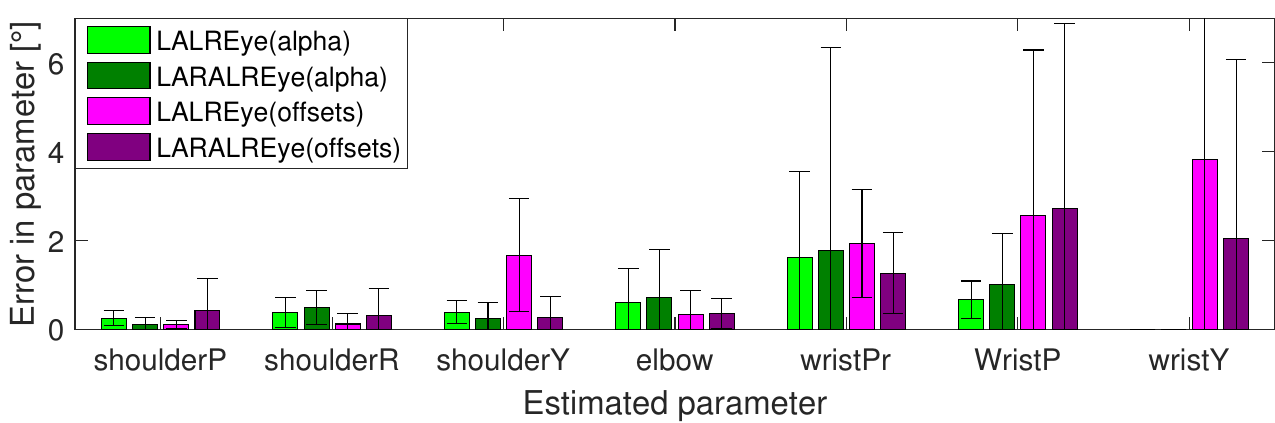}
    \caption{Absolute error of estimated DH parameters of LA chain after optimization (50 training poses, perturbation factor 5, measurement noise 5 px on cameras and 5 mm on touch). 
    (Top) $a$ and $d$ parameters. (Bottom) $\alpha$ and \textit{offsets}.}
    \label{fig:DHparametersEst}
\end{figure}

Finally, having showed above that the ``self-touch and self-observation'' (LARALREye) chain slightly outperforms the ``stereo self-observation'' only chain (LALREye) (Fig.~\ref{fig:pertDeg} top, Fig.~\ref{fig:MeasurementError}), also in observability (Fig.~ \ref{fig:observability}), here in Fig.~\ref{fig:DHparametersEst} we can observe a similar trend in the estimated parameters of the LA chain against their ground truth values. The parameter estimates obtained from LARALREye are significantly better for $d$ for all joints except for wristPr and elbow and for $a$ for all shoulder joints. The other parameters estimates are comparable. The wrist joint calibration seems to be sensitive on the selection of training poses and will need further study.

\section{Discussion and Conclusion}
We quantitatively and systematically investigated the potential of automatic self-contained kinematic calibration (DH parameters including camera extrinsic parameters) of a humanoid robot employing different kinematic chains---in particular relying on self-observation and self-touch. The parameters varied were: (i) type and number of intersecting kinematic chains used for calibration, (ii) parameters and chains subject to optimization, (iii) amount of initial perturbation of kinematic parameters, (iv) number of poses/configurations used for optimization, (v) amount of measurement noise in end-effector positions / cameras. We also tracked the computation time and while the details differ depending on the settings (chain calibrated, number of poses, etc.), a typical optimization run would not take more than tens of seconds on an older laptop PC.
Next to results w.r.t. the cost function itself (error on end-effector or camera reprojection) a number of additional analyses were performed including error residuals, errors on estimated parameters compared to ground truth, and observability analysis.

While some results were expected (such as improvement when more configurations are added or poor performance when using self-observation from a single camera), the most notable findings are: (1) calibrating parameters of a single chain (e.g. one arm) by employing multiple kinematic chains (``self-observation'' and ``self-touch'') is superior in terms of optimization results (Fig.~\ref{fig:pertDeg} top) as well as observability (Fig.~\ref{fig:observability}); (2) when using multi-chain calibration, fewer poses suffice to get similar performance compared to when e.g. only observation from a single camera is used  (Fig.~\ref{fig:pertDeg} top); (3) parameters of all chains (here 86 DH parameters) can be subject to calibration simultaneously and with 50 (100) poses, end-effector error of around 2 (1) mm can be achieved (Fig.~\ref{fig:pertDeg} bottom); (4) adding noise to a sensory modality degrades performance of all calibrations employing the chains relying on this information (Fig.~\ref{fig:MeasurementError}).
The last point is interesting to discuss in relation to Birbach et al. \cite{Birbach2015} who put forth the hypothesis that calibrating multiple chains simultaneously is superior to pairwise sequential calibration. Our results support this provided that measurement noise is small. Instead, if a certain modality is noisy, it may be beneficial to preferentially employ chains that rely on more accurate measurements first and then calibrate a ``noisy chain'' in a second step. 

We have only reported results from simulation, however, we claim that this was the right tool for this type of investigation. At the same time, our setup and choice of parameters was drawing on experiments performed in the real robot---self-touch \cite{Roncone_ICRA_2014} and self-observation \cite{Fanello2014,Vicente2016} in particular---which makes the results grounded in a real setting and should inform future experimentation on the iCub. The method to combine chains and analyze the results presented here can be transferred to other platforms as well. 

There are several aspects that we want to further investigate in the future. First, we note that while we did control for the angle between the palm and the contralateral finger for self-touch in the dataset generation, we did not monitor whether the contact point would be also visible. Additional analyses revealed that the contact point would not be occluded and hence be visible by both cameras in 35\% of the poses and by one of the cameras in 53\%. We recomputed the observability with this subset of the dataset only and found no decrease. In the future, configurations with occlusions should be excluded from dataset generation. Second, we found that around 50 configurations (data points) suffice for reasonable calibration. Finding the optimal subset of not more than 10 configurations would be desirable, such that recalibration can be performed rapidly. Here, clever pose selection will be necessary to warrant adequate and stable performance. Third, the information from the two cameras can be used to reproject observed position of the end-effector in image coordinates of both eyes (pixel $(u,v)$) to 3D space ($X^{eye}$) (similar to \cite{Fanello2014,Hirschmuller2008})---leading onto yet another formulation of the optimization problem.
Fourth, our investigation can be extended considering also the contribution of inertial sensors---in the robot head \cite{Birbach2015} or distributed on the robot body \cite{Guedelha2016,Mittendorfer2012}. Fifth, the present method can be compared with filtering approaches \cite{Vicente2016,Zenha2018} or with methods that pose fewer assumptions on the initial model available (e.g., \cite{Lanillos2018}).
Finally, the self-touch scenario can be also turned around from using a tactile array to calibrate kinematics \cite{Roncone_ICRA_2014,QiangLi2015} to calibrating the skin itself \cite{Albini2017}.

\section*{ACKNOWLEDGMENT}
 We thank Alessandro Roncone for assistance with the models of the iCub robot in MATLAB and the source files leading to Fig.~\ref{fig:kinModel} left and Ugo Pattacini for discussions, tips, and assistance with the use of Cartesian solvers leading to the generation of self-touch configurations.

%%%%%%%%%%%%%%%%%%%%%%%%%%%%%%%%%%%%%%%%%%%%%%%%%%%%%%%%%%%%%%%%%%%%%%%%%%%%%%%%
\bibliographystyle{IEEEtran}
\bibliography{selfcalib-multichain}

\end{document}